\documentclass[10pt,twocolumn,letterpaper]{article}

\usepackage{iccv}
\usepackage{times}
\usepackage{epsfig}
\usepackage{graphicx}
\usepackage{amsmath}
\usepackage{amssymb}
\usepackage{booktabs} 
\usepackage{multirow}
\usepackage{bbm}

\usepackage[breaklinks=true,bookmarks=false]{hyperref}

\iccvfinalcopy 

\def\iccvPaperID{3992} 

\ificcvfinal\fi
\begin{document}

\title{Self-Ensembling with GAN-based Data Augmentation for Domain Adaptation in Semantic Segmentation}

\author{Jaehoon Choi\\
KAIST\\
\and
Taekyung Kim\\
KAIST\\
\and
Changick Kim\\
KAIST\\
\and
{\tt\small \{whdns44, tkkim93, changick\}@kaist.ac.kr}
}

\maketitle
\ificcvfinal\thispagestyle{empty}\fi

\begin{abstract}
Deep learning-based semantic segmentation methods have an intrinsic limitation that training a model requires a large amount of data with pixel-level annotations. To address this challenging issue, many researchers give attention to unsupervised domain adaptation for semantic segmentation. Unsupervised domain adaptation seeks to adapt the model trained on the source domain to the target domain. In this paper, we introduce a self-ensembling technique, one of the successful methods for domain adaptation in classification. However, applying self-ensembling to semantic segmentation is very difficult because heavily-tuned manual data augmentation used in self-ensembling is not useful to reduce the large domain gap in the semantic segmentation. To overcome this limitation, we propose a novel framework consisting of two components, which are complementary to each other. First, we present a data augmentation method based on Generative Adversarial Networks (GANs), which is computationally efficient and effective to facilitate domain alignment. Given those augmented images, we apply self-ensembling to enhance the performance of the segmentation network on the target domain. The proposed method outperforms state-of-the-art semantic segmentation methods on unsupervised domain adaptation benchmarks.  
\end{abstract}

\section{Introduction}
Semantic segmentation has been widely studied in the computer vision field. Its goal is to assign image category labels to each pixel in the image. A wide variety of algorithms based on deep neural networks have achieved high performance with sufficient amounts of annotated datasets. However, creating large labeled datasets for semantic segmentation is cost-expensive and time-consuming \cite{cordts2016cityscapes}. To overcome the annotation burden, researchers utilize modern computer graphics to easily generate synthetic images with ground truth labels~\cite{richter2016playing}. Unfortunately, in practice, models trained with synthetic data do not perform well on a realistic domain because there exists a distribution difference called domain shift. Unsupervised domain adaptation handles the domain shift by transferring knowledge from the labeled dataset in the source domain to the unlabeled dataset in the target domain~\cite{ben2010theory}. 

\begin{figure}[t]
\begin{center}
\includegraphics[width=0.95\linewidth]{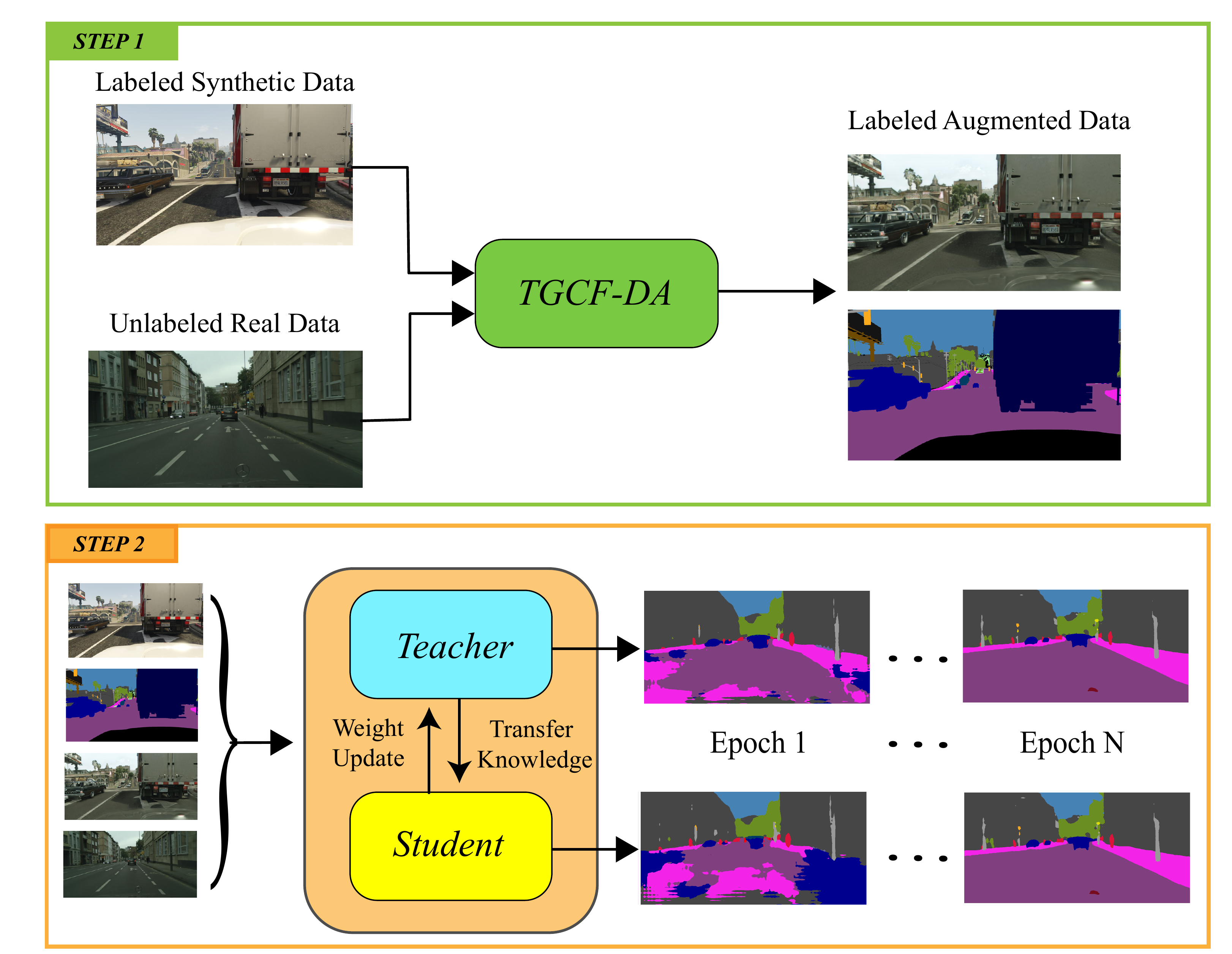}
\end{center}
   \caption{The overall framework of our method. Given labeled synthetic data and unlabeled real data, we propose a Target-Guided and Cycle-Free Data Augmentation (TGCF-DA) method to generate labeled augmented data (green). We introduce two segmentation networks as the teacher and the student in order to implement the self-ensembling algorithm (orange). Both segmentation networks are trained by augmented data as well as synthetic and real data. During the learning process, the teacher network transfers its knowledge to the student network.}
\label{fig:introduction}
\end{figure}

 Recent approaches for domain adaptation focus on aligning features extracted from the source and target data. In particular, most of the domain adaptation methods in semantic segmentation depend on adversarial training aiming to minimize the domain discrepancy through domain confusion \cite{hoffman2016fcns, pmlr-v80-hoffman18a, tsai2018learning, sankaranarayanan2018learning, hong2018conditional,zhang2018fully}. However, adversarial approaches suffer from a significant drawback. Since these methods seek to align the global distributions of two different domains, the adversarial loss may trigger a negative transfer, which aligns the target feature with the source feature in an incorrect semantic category. The negative transfer can have adverse effect on features that are already well aligned. Thus, this adaptation often performs even worse than a network trained solely on the source domain. Instead of adversarial training, we take an alternative way to perform feature-level domain alignment. We adopt self-ensembling \cite{french2017self}, one of the effective methods for domain adaptation in classification.  

  Self-ensembling is composed of a teacher and a student network, where the student is compelled to produce consistent predictions provided by the teacher on target data. As the teacher is an ensembled model that averages the student's weights, predictions from the teacher on target data can be thought of as the pseudo labels for the student. While recent self-ensembling proves its effectiveness in classification, these approaches require heavily-tuned manual data augmentation \cite{french2017self} for successful domain alignment. Furthermore, although such data augmentation consisting of various geometric transformations is effective in classification, it is not suited to minimize the domain shift in semantic segmentation. Two different geometric transformations on each input can cause spatial misalignment between the student and teacher predictions. Thus, we propose a novel data augmentation method to deal with this issue.

Our augmented image synthesis method is based on generative adversarial networks (GANs) \cite{goodfellow2014generative}.  We aim to generate augmented images, in which semantic contents are preserved, because these images with the inconsistent semantic content impair the segmentation performance due to the pixel-level misalignment between augmented images and source labels. Hence, we add a semantic constraint for the generator to preserve global and local structures, \ie the semantic consistency. Furthermore, we propose a target-guided generator, which produces images conditioned on style information extracted from the target domain. In other words, our generator synthesizes augmented images maintaining semantic information, while only transferring styles from target images. 

Most previous studies for GAN-based Image-to-Image translation methods \cite{zhu2017unpaired, yi2017dualgan, liu2017unsupervised, lee2018diverse, huang2018multimodal, huang2018auggan, ma2018exemplar} rely on various forms of cycle-consistency. However, incorporating cycle-consistency into unsupervised domain adaptation has two limitations. First, it needs redundant modules such as a target-to-source generator and corresponding computational burden. Second, cycle-consistency may be too strong when target data are scarce compared to source data \cite{hosseini2018augmented}, which is the general setting of unsupervised domain adaptation. Our proposed model does not consider all kinds of cycle-consistency. We refer to our method as Target-Guided and Cycle-Free Data Augmentation (TGCF-DA).

Our universal framework is illustrated in Fig.\:\ref{fig:introduction}. We employ TGCF-DA to produce augmented images. Then, the segmentation network learns from the source, target and augmented data through self-ensembling. The main contributions of this paper are summarized as follows: 
\begin{itemize}
    \item We propose a novel data augmentation method with a target-guided generator and a cycle-free loss which is more efficient and suitable for semantic segmentation in unsupervised domain adaptation.    
    \item We build a unified framework that collaborates the self-ensembling with TGCF-DA.
    \item Our approach achieves the state-of-the-art performances on challenging benchmark datasets. Also, we conduct extensive experiments and provide comprehensive analyses for the proposed method.
\end{itemize}

\section{Related work}
\textbf{Unsupervised Domain Adaptation for Semantic Segmentation:}
Recently unsupervised domain adaptation for semantic segmentation has received much attention. The first attempt to this task is FCNs in the wild \cite{hoffman2016fcns}, which simultaneously performs the global and local alignment with adversarial training. Adversarial training is the predominant approach focusing on a feature-level adaptation to generate domain-invariant features through domain confusion, e.g., \cite{chen2017no, chen2018road, tsai2018learning, sankaranarayanan2018learning, hong2018conditional, saito2018maximum, huang2018domain}. This idea is extended to jointly adapt representations at both pixel and feature level through various techniques such as cycle-consistency loss \cite{pmlr-v80-hoffman18a, murez2018image} or style transfer \cite{dundar2018domain, wu2018dcan}. Except for adversarial training methods, there is a different approach based on self-training. CBST \cite{zou2018unsupervised} introduces self-training to produce pseudo labels and retrain the network with these labels. 

\textbf{Self-Ensembling:} Self-Ensembling \cite{zhu2005semi,Rosenberg:2005:SSO:1042449.1043907} is proposed in the field of semi-supervised learning. A popular method for semi-supervised learning is the consistency regularization, which employs unlabeled data to produce consistent predictions under perturbations \cite{sajjadi2016regularization, athiwaratkun2018there}. Laine and Aila \cite{laine2016temporal} propose Temporal Ensembling using a per-sample moving average of predictions for the consistent output. Tarvainen and Valpola \cite{tarvainen2017mean} suggest an exponential moving average of the model weights instead of average of predictions. The self-ensembling method \cite{french2017self} applies a Mean Teacher framework to unsupervised domain adaptation with some modifications. In \cite{perone2018unsupervised}, Perone \etal address medical imaging segmentation tasks by applying the self-ensembling method akin to the previous method. Yonghao \etal \cite{SEAN} utilize the self-ensembling attention network to extract attention-aware features for domain adaptation.    

\begin{figure*}[t]
\begin{center}
\includegraphics[width=0.9\linewidth]{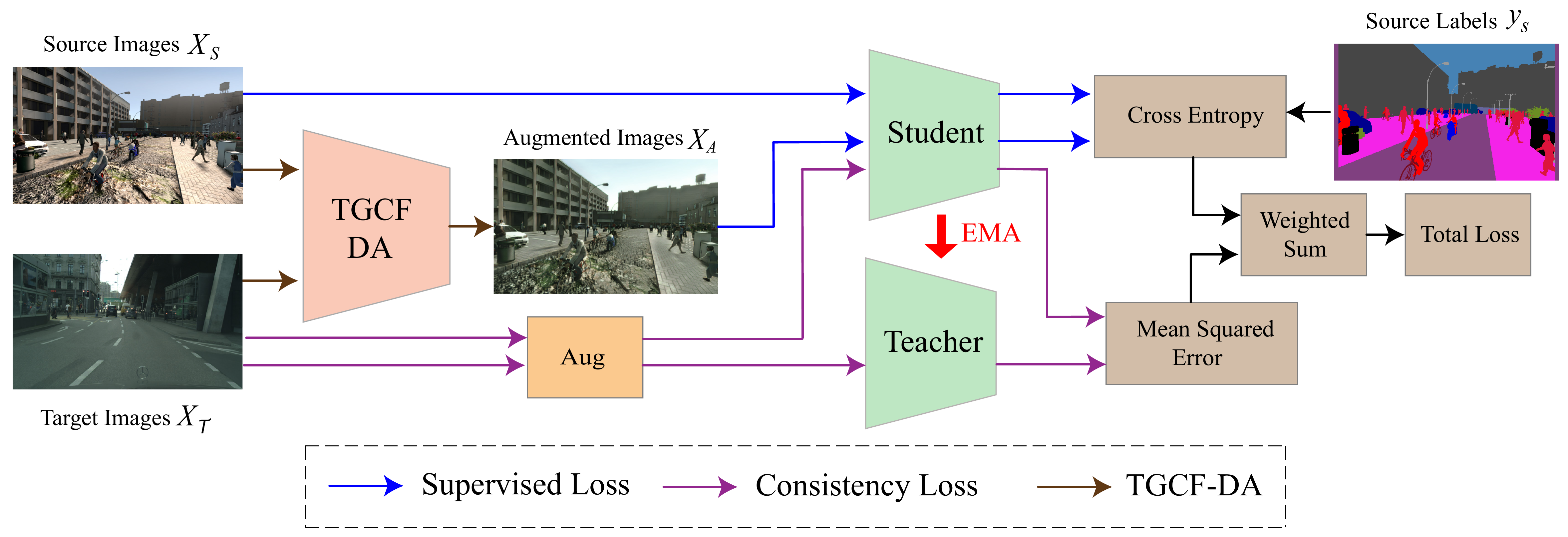}
\end{center}
   \caption{An overview of the proposed framework. 1) The source and target images are fed into Target-Guided generator to produce augmented images. 2) The supervised loss is a multi-class cross-entropy loss with source images and augmented images. 3) The consistency loss is a mean squared error between both prediction maps extracted from the student and teacher network. 4) A total training loss is the weighted sum of the supervised loss and the consistency loss. 5) We perform data augmentation only for target samples to complement the consistency loss. 6) The teacher network's weights are the exponential moving average (EMA) of those of the student network.}
\label{fig_main}
\end{figure*}

\textbf{Image-to-Image Translation:} Recent approaches for Image-to-Image (I2I) Translation are based on Generative Adversarial Networks (GANs) \cite{goodfellow2014generative}. In the case of unpaired training images, one popular constraint is cycle-consistency that maps a given image to the target domain and reconstructs the original image \cite{kim2017learning,zhu2017unpaired, yi2017dualgan}. UNIT \cite{liu2017unsupervised} introduces a constraint for learning a shared latent space. However, all the aforementioned methods suffer from a lack of diversity in translated images. To produce multi-modal outputs, one possible approach injects noise vectors as additional inputs to the generator \cite{zhu2017toward, almahairi2018augmented, gonzalez2018image}, but it could lead to the mode collapse problem. Also, Since cycle-consistency is too restrictive, variants of cycle-consistency \cite{zhu2017toward, huang2018multimodal, lee2018diverse} are developed for multi-modal I2I translation. A different approach is to apply neural style transfer \cite{gatys2016image,ulyanov2016instance,johnson2016perceptual,huang2017arbitrary}. In particular, concurrent works \cite{huang2018multimodal, ma2018exemplar} employ an adaptive instance normalization \cite{huang2017arbitrary} to transfer style from the exemplar to the original image. In addition, the authors of AugGAN \cite{huang2018auggan} exploit the segmentation information for improving I2I translation network. Our task is entirely different from AugGAN because domain adaptation cannot use segmentation labels of the target data.         

\section{Proposed Method}
 In this work, we introduce the unified framework, which is built upon the self-ensembling for semantic segmentation. The key to improve the capacity of the self-ensembling for semantic segmentation is the GAN-based data augmentation to align representations of source and target rather than geometric transformations mostly used in existing self-ensembling for classification. To achieve this goal, we present a novel Target-Guided and Cycle-Free Data Augmentation (TGCF-DA) with a target-guided generator and a semantic constraint. The target-guided generator translates source images to different styles in the target domain. Our student network learns the source images and augmented images from TGCF-DA with a supervised loss by computing cross-entropy loss. Also, we only use target samples to compute the consistency loss, which is defined as the mean squared error between prediction maps generated from the student and teacher networks.
 
 More formally, let \(X_{\mathcal{S}}\) and \(X_{\mathcal{T}}\) denote the source domain and target domain. We have access to \(N_{s}\) labeled source samples \(\{(x_{s}^i, y_{s}^i)\}_{i=1}^{N_{s}}\) with \(x_{s}^i\in X_{\mathcal{S}}\) and the corresponding label maps \(y_{s}^i\). The target domain has \(N_{t}\) unlabeled target samples \(\{x_{t}^i\}_{i=1}^{N_{t}}\), where \(x_{t}^i \in X_{\mathcal{T}}\). \(P_{\mathcal{S}}\) and \(P_{\mathcal{T}}\) denote the source and target data distributions, respectively. The source and target data share \(C\) categories. Let \(f_{S}\) and \(f_{T}\) be a student segmentation network and a teacher segmentation network.   

\subsection{Target-guided generator}
\label{section Target-guided generator}
 Based on the assumption that image can be decomposed into two disentangled representations \cite{liu2017unsupervised, huang2018multimodal}, a content and a style, we adopt a source encoder for generating content representation and a target encoder for extracting style representation. To combine these two representations properly, we apply Adaptive Instance Normalization (AdaIN) \cite{huang2017arbitrary} to feature maps of source images. As in \cite{huang2018multimodal}, the target encoder with multiple fully connected layers provide the learnable affine transformation parameters \((\gamma_{t}, \beta_{t})\) to normalize the feature maps of a source image for each channel. The AdaIN operation is defined as:  
\begin{equation} \label{AdaIN}
\Tilde{F}_{s}^i = \gamma_{t}^{i}\:(\frac{F_{s}^i - \mu(F_{s}^i)}{\sigma(F_{s}^i)}) + \beta_{t}^{i}\:,
\end{equation}
where \(F_{s}^i\) denotes the source feature map for the \(i\)-th channel.\:\:\(\mu(\cdot)\) and \(\sigma(\cdot)\) respectively denote mean and variance across spatial dimensions. Our generator is guided by the style information of target samples through AdaINs at intermediate residual blocks while preserving the spatial structure of source images, \ie the semantic consistency of source images is retained.

\begin{figure}[t]
\begin{center}
\includegraphics[width=0.95\linewidth]{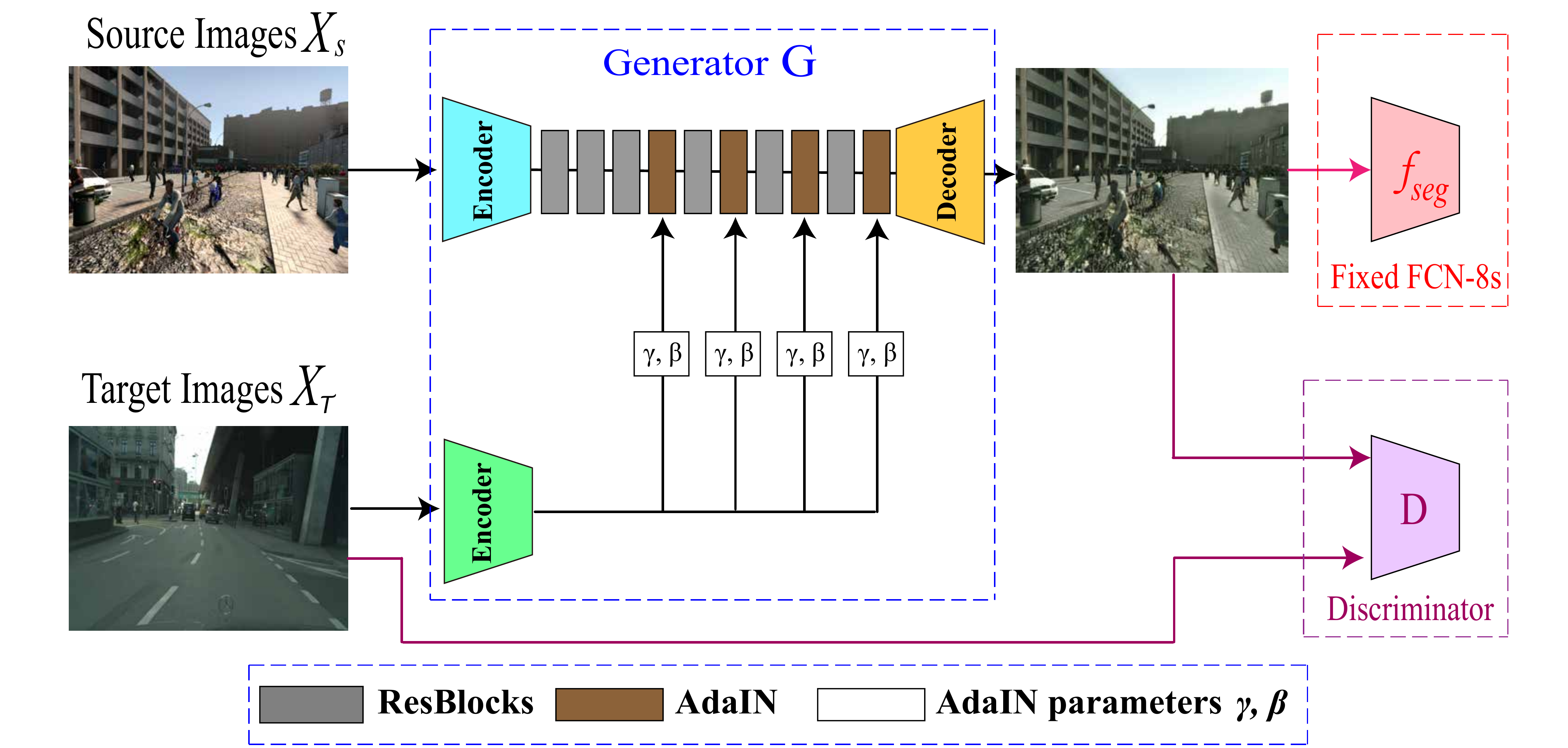}
\end{center}
   \caption{The overview of TGCF-DA based on GAN \cite{goodfellow2014generative}. The blue box describes the target-guided generator \(G\). The red box is the pretrained segmentaion model \(f_{seg}\) with fixed weights. The purple box is the discriminator \(D\).}
\label{fig_target_guided_generator}
\end{figure}

\subsection{Semantic constraint}
\label{section semantic constraint}
    We utilize a semantic constraint to preserve semantic content at pixel level. Given the labeled source data, we can pretrain the segmentation model such as FCN-8s \cite{long2015fully} for constraining the generator. The pretrained segmentation model \(f_{seg}\) with fixed weights encourages the semantic consistency between the images before and after the translation. Thanks to this semantic constraint, our network can preserve the objects in images without distortion. Furthermore, this constraint is crucial to stabilizing the adversarial training without the cycle-consistency. Since the cycle-consistency enforces the strict constraint for matching two distributions, it is effective to prevent the mode collapse and to stabilize the adversarial training \cite{li2017alice}. Without the cycle-consistency, our adversarial training is vulnerable to instability of GAN training. However, this semantic constraint guarantees stable adversarial training by strongly enforcing semantic consistency. We define the semantic constraint loss as the cross-entropy loss:
\begin{equation} \label{Semantic_Constraint}
\begin{split}
L_{sem}&(f_{seg}, G) =\\
&\frac{-1}{HW}\sum_{k=1}^{H \times W} \sum_{c=1}^{C} y_{s}^{(k,c)} \:log(f_{seg}(G(x_{s}, x_{t}))^{(k, c)})\:, \\
\end{split}
\end{equation}
where \(G(x_{s}, x_{t})\) is the generated image of size \(H\)\(\times\)\(W\) produced by the target-guided generator G.

\subsection{Target-guided and cycle-free data augmentation}
 We introduce a GAN designed for Target-Guided and Cycle-Free Data Augmentation (TGCF-DA). As in Fig.\:\ref{fig_target_guided_generator}, \(G\) is the target-guided generator and \(D\) is the discriminator proposed in \cite{wang2018high}. We use the adversarial objective from LSGAN \cite{mao2017least} and apply the spectral normalization \cite{miyato2018spectral} to stabilize the GAN training. The GAN loss is defined as:
\begin{equation} \label{TGCF-DA_GAN_Loss}
\begin{split}
L_{GAN}(G, D) &= E_{(x_{s}, x_{t}) \sim (P_{\mathcal{S}}, P_{\mathcal{T}})}[D(G(x_{s}, x_{t}))^2] \\
&+E_{x_{t} \sim P_{\mathcal{T}}}[(D(x_{t})-1)^2]\:. \\
\end{split}
\end{equation}
This loss ensures that \(G\) produces new images visually similar to target images without losing semantic content in source images. Since the segmentation model \(f_{seg}\) is fixed, we jointly train the target-guided generator and discriminator to optimize the overall loss:
\begin{equation} \label{TGCF-DA_Overall_Loss}
\begin{split}
L_{TGCF-DA} = L_{GAN} +\:\lambda_{sem}L_{sem}\:,
\end{split}
\end{equation}
where \(\lambda_{sem}\) is a weight to balance the contribution of the GAN loss and semantic constraint. The pretrained target-guided generator is employed to synthesize augmented images with the purpose of data augmentation in the self-ensembling.  


\subsection{Self-ensembling} \label{section self-ensembling}
We construct the teacher network \(f_{T}\) and the student network \(f_{S}\). The teacher's weights \(t_{i}\) at training step \(i\) are updated by the student's weights \(s_{i}\) following the formula:
\begin{equation} \label{weight_ema}
t_{i} = \alpha t_{i-1} + (1 - \alpha)s_{i}\:,    
\end{equation}
where \(\alpha\) is an exponential moving average decay. During training, each mini-batch consists of source samples, augmented samples, and target samples. We use source samples and augmented samples to compute the supervised loss \(L_{sup}\), which is cross-entropy function for semantic segmentation. This loss function enables the student network to produce the semantically accurate prediction for the source and augmented samples. The consistency loss \(L_{con}\) is formulated as the mean-squared error between the prediction maps generated from the student and teacher network:
\begin{equation} \label{MSE}
\begin{split}
L_{con}(f_{S}, f_{T}) = E_{x_{t} \sim P_{\mathcal{T}}}[\|\sigma(f_{S}(x_{t})) - \sigma(f_{T}(x_{t}))\|^2]\:,
\end{split}
\end{equation}
where \(\sigma\) is a softmax function to compute probability of prediction maps. The total loss \(L_{total}\) is the weighted sum of the supervised loss \(L_{sup}\) and the consistency loss \(L_{con}\):
\begin{equation} \label{totalloss}
\begin{split}
L_{total} = L_{sup} + \delta_{con} L_{con}\:, 
\end{split}
\end{equation}
where \(\delta_{con}\) is the weight of consistency loss subject to the ramp-ups. Contrary to \cite{french2017self}, we empirically observe that weight ramp-up is necessary for enhancing the effectiveness of the consistency loss. 

\subsection{Data augmentation for target samples}
 Here, data augmentation for target samples is not relevant to TGCF-DA. This data augmentation is only applied to target samples in order to compute the consistency loss for the self-ensembling in Section \ref{section self-ensembling}. In classification \cite{french2017self}, the goal of random data augmentations for target samples is forcing the student network to produce different predictions for the same target sample. Aforementioned above, image-level transformations such as geometric transformations are not helpful for the pixel-level prediction task like semantic segmentation \cite{liu2018pixel}. Thus, we inject Gaussian noise to target samples, which are fed to student and target networks respectively. In addition, we apply Dropout \cite{srivastava2014dropout} for weight perturbation. As a result, our student network is forced to produce consistent predictions with the teacher network under different perturbations for target samples and parameters of each network. 

\section{Experiments}
This section describes experimental setups and details of synthetic-to-real domain adaptation. Then, we will report the experiment results compared with the previous researches. Furthermore, we will provide ablation studies to validate the effectiveness of our method. 

\subsection{Datasets}
For a synthetic source domain, we used SYNTHIA \cite{ros2016synthia} and GTA5 \cite{richter2016playing} datasets. Then, we evaluated our method on Cityscapes dataset \cite{cordts2016cityscapes} as a real-world target domain following similar settings in \cite{hoffman2016fcns, zhang2017curriculum, tsai2018learning, sankaranarayanan2018learning}. We briefly introduce the details of datasets as following:  

\noindent\textbf{GTA5.} GTA5 \cite{richter2016playing} contains 24966 urban scene images with pixel-level annotations. These high-resolution images are rendered from the gaming engine Grand Theft Auto V. Following \cite{hoffman2016fcns}, we used the 19 categories of the annotations compatible with those of the Cityscapes. We randomly picked 1000 images from GTA5 for validation purpose.

\noindent\textbf{SYNTHIA.} SYNTHIA \cite{ros2016synthia} is a large-scale dataset of video sequences rendered from a virtual city. We used SYNTHIA-RAND-CITYSCAPES, consisting of 9400 images with pixel-level annotations. Inheriting from the previous work \cite{zhang2017curriculum}, we chose 16 categories common in both SYNTHIA and Cityscapes. We randomly selected 100 images for evaluation.

\noindent\textbf{Cityscapes.} Cityscapes \cite{cordts2016cityscapes} contains urban street scenes collected from 50 cities around Germany and neighboring countries. It has a training set with 2975 images and a validation set with 500 images.

We can utilize source images and labels from either SYNTHIA or GTA5, as well as target images without labels from the training set of Cityscapes. The validation set in Cityscapes is treated as the evaluation set for our domain adaptation experiment. We report IoU (Intersection-over-Union) for each class and mIoU (mean IoU) to measure the segmentation performance. In supplementary material, we provide additional experimental results on the BDD100K dataset \cite{yu2018bdd100k}.        

\subsection{Experiment setup and implementation details}
\noindent\textbf{TGCF-DA.} Our augmentation network for TGCF-DA is composed of the generator, the discriminator and the segmentation model. The generator is built upon the auto-encoder architecture used by MUNIT \cite{huang2018multimodal}, but modified to act as the cycle-free generator. It consists of the source encoder, the target encoder and the decoder. The source encoder includes strided convolutional layers to downsample the source images and residual blocks \cite{he2016deep} to compute the content representations. The decoder consists of residual blocks and transposed convolutional layers to upsample the combined representations. The target encoder is comprised of strided convolutional layers and fully connected layers to provide the style representations. Multi-scale discriminators described in \cite{wang2018high} are employed as our discriminator. We set the weight \(\lambda_{sem}\) to 10 in all experiments. 

\noindent\textbf{Self-Ensembling.} In all our experiments, we employed a VGG-16 backbone for our semantic segmentation network. Following Deeplab \cite{chen2018deeplab}, we incorporated ASPP (Atrous Spatial Pyramid Pooling) as the decoder and then used an upsampling layer to get the final segmentation output. Before the upsampling layer, the output of the final classifier is used to compute the consistency loss in Section \ref{section self-ensembling}. Motivated by \cite{tarvainen2017mean}, we utilized the sigmoid ramp-up for the consistency loss weight \(\delta_{con}\). The details of the consistency loss weight is analyzed in Section \ref{Subsection:Hyperparameter sensitivity on self-ensembling}. During training process, the images are resized and cropped to 480\(\times\)960 resolution, and for evaluation we upsample our prediction maps to 1024\(\times\)2048 resolution.  The details of our architecture and experiments will be available in the supplementary material.

\subsection{Experimental results}
\begin{table*}[h]
\begin{center}
\begin{small}
\resizebox{\textwidth}{!}{\begin{tabular}{l||c||ccccccccccccccccccc||c}
\toprule
\multicolumn{22}{c}{(a) \textbf{GTA5} \(\rightarrow\) \textbf{Cityscapes}}\\
\hline
\textup{Method} & \rotatebox[origin=c]{90}{Mech.} & \rotatebox[origin=c]{90}{road} & \rotatebox[origin=c]{90}{sidewalk} & \rotatebox[origin=c]{90}{building} & \rotatebox[origin=c]{90}{wall} & \rotatebox[origin=c]{90}{fence} & \rotatebox[origin=c]{90}{pole} & \rotatebox[origin=c]{90}{light} & \rotatebox[origin=c]{90}{sign} & \rotatebox[origin=c]{90}{vegetation} & \rotatebox[origin=c]{90}{terrain} & \rotatebox[origin=c]{90}{sky} & 
\rotatebox[origin=c]{90}{person} & \rotatebox[origin=c]{90}{rider} &
\rotatebox[origin=c]{90}{car} & \rotatebox[origin=c]{90}{truck} & 
\rotatebox[origin=c]{90}{bus} & \rotatebox[origin=c]{90}{train}& 
\rotatebox[origin=c]{90}{motorbike} & \rotatebox[origin=c]{90}{bike} & \rotatebox[origin=c]{90}{\textbf{mIoU}}\\
\hline
Baseline (Source Only) & - & 61.0 & 18.5 & 66.2 & 18.0 & 19.6 & 19.1 & 22.4 & 15.5 & 79.6 & 28.5 & 58.0 & 44.5 & 1.7 & 66.6 & 14.1 & 1.1 & 0.0 & 3.2 & 0.7 & 28.3 \\
\hline
\hline
Curriculum DA \cite{zhang2017curriculum} & ST & 72.9 & 30.0 & 74.9 & 12.1 & 13.2 & 15.3 & 16.8 & 14.1 & 79.3 & 14.5 & 75.5 & 35.7 & 10.0 & 62.1 & 20.6 & 19.0 & 0.0 & 19.3 & 12.0 & 31.4 \\
\hline
CyCADA \cite{pmlr-v80-hoffman18a} & AT & 85.2 & 37.2 & 76.5 & 21.8 & 15.0 & 23.8 & 22.9 & 21.5 & 80.5 & 31.3 & 60.7 & 50.5 & 9.0 & 76.9 & 17.1 & 28.2 & 4.5 & 9.8 & 0.0 & 35.4 \\
\hline
MCD \cite{saito2018maximum} & AT & 86.4 & 8.5 & 76.1 & 18.6 & 9.7 & 14.9 & 7.8 & 0.6 & 82.8 & 32.7 & 71.4 & 25.2 & 1.1 & 76.3 & 16.1 & 17.1 & 1.4 & 0.2 & 0.0 & 28.8 \\
\hline
LSD-seg \cite{sankaranarayanan2018learning} & AT & 88.0 & 30.5 & 78.6 & 25.2 & \textbf{23.5} & 16.7 & 23.5 & 11.6 & 78.7 & 27.2 & 71.9 & 51.3 & 19.5 & 80.4 & 19.8 & 18.3 & 0.9 & 20.8 & 18.4 & 37.1 \\
\hline
AdaptSegNet \cite{tsai2018learning} & AT & 87.3 & 29.8 & 78.6 & 21.1 & 18.2 & 22.5 & 21.5 & 11.0 & 79.7 & 29.6 & 71.3 & 46.8 & 6.5 & 80.1 & 23.0 & 26.9 & 0.0 & 10.6 & 0.3 & 35.0 \\
\hline
ROAD \cite{chen2018road} & AT & 85.4 & 31.2 & 78.6 & \textbf{27.9} & 22.2 & 21.9 & 23.7 & 11.4 & 80.7 & 29.3 & 68.9 & 48.5 & 14.1 & 78.0 & 19.1 & 23.8 & \textbf{9.4} & 8.3 & 0.0 & 35.9 \\
\hline
Conservative Loss \cite{zhu2018penalizing} & AT & 85.6 & 38.3 & 78.6 & 27.2 & 18.4 & 25.3 & 25.0 & 17.1 & 81.5 & 31.3 & 70.6 & 50.5 & \textbf{22.3} & 81.3 & 25.5 & 21.0 & 0.1 & 18.9 & 4.3 & 38.1 \\
\hline
DCAN \cite{wu2018dcan} & SR & 82.3 & 26.7 & 77.4 & 23.7 & 20.5 & 20.4 & 30.3 & 15.9 & 80.9 & 25.4 & 69.5 & 52.6 & 11.1 & 79.6 & 24.9 & 21.2 & 1.3 & 17.0 & 6.7 & 36.2 \\
\hline
CBST \cite{zou2018unsupervised} & ST & 66.7 & 26.8 & 73.7 & 14.8 & 9.5 & 28.3 & 25.9 & 10.1 & 75.5 & 15.7 & 51.6 & 47.2 & 6.2 & 71.9 & 3.7 & 2.2 & 5.4 & 18.9 & \textbf{32.4} & 30.9 \\
\hline
Self-Ensembling (SE) & ST & 76.4 & 16.7 & 71.5 & 13.0 & 13.1 & 17.5 & 17.3 & 8.3 & 76.5 & 16.3 & 67.4 & 42.5 & 10.4 & 78.1 & \textbf{27.9} & \textbf{37.2} & 0.0 & 22.2 & 7.4 & 32.6\\
TGCF-DA & AT & 73.9 & 19.8 & 74.8 & 19.7 & 21.8 & 20.7 & 26.7 & 12.4 & 78.0 & 22.3 & 72.0 & 53.4 & 12.9 & 73.3 & 24.5 & 28.5 & 0.0 & \textbf{24.3} & 14.1 & 35.4 \\
Ours (TGCF-DA + SE) & AT+ST & \textbf{90.2} & \textbf{51.5} & \textbf{81.1} & 15.0 & 10.7 & \textbf{37.5} & \textbf{35.2} & \textbf{28.9} & \textbf{84.1} & \textbf{32.7} & \textbf{75.9} & \textbf{62.7} & 19.9 & \textbf{82.6} & 22.9 & 28.3 & 0.0 & 23.0 & 25.4 & \textbf{42.5} \\
\hline
\hline
Target Only & - & 94.3 & 77.7 & 86.6 & 52.9 & 50.4 & 50.1 & 52.9 & 57.0 & 81.4 & 64.8 & 94.1 & 57.8 & 55.5 & 87.6 & 79.0 & 56.1 & 19.6 & 45.3 & 20.9 & 62.3  \\
\bottomrule
\end{tabular}}
\end{small}
\end{center}

\begin{center}
\begin{small}
\resizebox{\textwidth}{!}{\begin{tabular}{l||c||cccccccccccccccc||c|c}
\toprule
\multicolumn{20}{c}{(b) \textbf{SYNTHIA} \(\rightarrow\) \textbf{Cityscapes}}\\
\hline
\textup{Method} & \rotatebox[origin=c]{90}{Mech.} & \rotatebox[origin=c]{90}{road} & \rotatebox[origin=c]{90}{sidewalk} & \rotatebox[origin=c]{90}{building} & \rotatebox[origin=c]{90}{wall} & \rotatebox[origin=c]{90}{fence} & \rotatebox[origin=c]{90}{pole} & \rotatebox[origin=c]{90}{light} & \rotatebox[origin=c]{90}{sign} & \rotatebox[origin=c]{90}{vegetation} & \rotatebox[origin=c]{90}{sky} & 
\rotatebox[origin=c]{90}{person} & \rotatebox[origin=c]{90}{rider} &
\rotatebox[origin=c]{90}{car} & \rotatebox[origin=c]{90}{bus} & 
\rotatebox[origin=c]{90}{motorbike} & \rotatebox[origin=c]{90}{bike} & \rotatebox[origin=c]{90}{\textbf{mIoU}} & \rotatebox[origin=c]{90}{\textbf{mIoU*}} \\ 
\hline
Baseline (Source Only) & -& 6.8 & 15.4 & 56.8 & 0.8 & 0.1 & 14.6 & 4.7 & 6.8 & 72.5 & 78.6 & 41.0 & 7.8 & 46.9 & 4.7 & 1.8 & 2.1 & 22.6 & 24.1 \\ 
\hline
\hline
Curriculum DA \cite{zhang2017curriculum} & ST & 65.2 & 26.1 & 74.9 & 0.1 & 0.5 & 10.7 & 3.7 & 3.0 & 76.1 & 70.6 & 47.1 & 8.2 & 43.2 & 20.7 & 0.7 & 13.1 &  29.0 & 34.8 \\
\hline
LSD-seg \cite{sankaranarayanan2018learning} & AT & 80.1 & 29.1 & 77.5 & 2.8 & 0.4 & 26.8 & 11.1 & 18.0 & 78.1 & 76.7 & 48.2 & 15.2 & 70.5 & 17.4 & 8.7 & 16.7 & 36.1 & - \\
\hline
AdaptSegNet \cite{tsai2018learning} & AT & 78.9 & 29.2 & 75.5 & - & - & - & 0.1 & 4.8 & 72.6 & 76.7 & 43.4 & 8.8 & 71.1 & 16.0 & 3.6 & 8.4 & - & 37.6 \\
\hline
ROAD \cite{chen2018road} & AT & 77.7 & 30.0 & 77.5 & 9.6 & 0.3 & 25.8 & 10.3 & 15.6 & 77.6 & 79.8 & 44.5 & 16.6 & 67.8 & 14.5 & 7.0 & 23.8 & 36.2 & - \\
\hline
Conservative Loss \cite{zhu2018penalizing} & AT & 80.0 & 31.4 & 72.9 & 0.4 & 0.0 & 22.4 & 8.1 & 16.7 & 74.8 & 72.2 & 50.9 & 12.7 & 53.9 & 15.6 & 1.7 & 33.5 & 34.2 & 40.3 \\
\hline
DCAN \cite{wu2018dcan}& SR & 79.9 & 30.4 & 70.8 & 1.6 & \textbf{0.6} & 22.3 & 6.7 & \textbf{23.0} & 76.9 & 73.9 & 41.9 & \textbf{16.7} & 61.7 & 11.5 & \textbf{10.3} & \textbf{38.6} & 35.4 & - \\
\hline
CBST \cite{zou2018unsupervised} & ST & 69.6 & 28.7 & 69.5 & \textbf{12.1} & 0.1 & 25.4 & \textbf{11.9} & 13.6 & 82.0 & \textbf{81.9} & 49.1 & 14.5 & 66.0 & 6.6 & 3.7 & 32.4 & 35.4 & 36.1 \\
\hline
Self-Ensembling (SE) & ST & 40.1 & 19.6 & 75.2 & 2.6 & 0.2 & 23.2 & 4.0 & 9.8 & 60.3 & 38.3 & 49.1 & 14.0 & 67.0 & 17.4 & 6.4 & 11.9 & 27.5 & 29.2  \\
TGCF-DA & AT & 63.9 & 25.6 & 75.9 & 5.4 & 0.1 & 22.6 & 2.6 & 6.8 & 78.4 & 77.2 & 48.7 & 16.5 & 62.2 & \textbf{24.2} & 5.0 & 22.1 & 33.6 & 39.8 \\
Ours (TGCF-DA + SE) & AT+ST & \textbf{90.1} & \textbf{48.6} & \textbf{80.7} & 2.2 & 0.2 & \textbf{27.2} & 3.2 & 14.3 & \textbf{82.1} & 78.4 & \textbf{54.4} & 16.4 & \textbf{82.5} & 12.3 & 1.7 & 21.8 & \textbf{38.5} & \textbf{46.6} \\
\hline
\hline
Target Only & - & 89.2 & 85.3 & 90.7 & 65.5 & 60.7 & 21.5 & 2.1 & 7.2 & 74.2 & 93.2 & 61.8 & 40.1 & 78.4 & 81.4 & 36.7 & 24.8 & 57.1 & 64.1 \\
\bottomrule
\end{tabular}}
\end{small}
\end{center}
\caption{The semantic segmentation results on Cityscapes validation set when evaluating the model trained on (a) GTA5 and (b) SYNTHIA. All segmentation models in table use VGG-16 based models. The mIoU* denotes the segmentation results over the 13 common classes. \textquotedblleft Source Only" denotes the evaluation result of models only trained on source data. \textquotedblleft  Target Only" denotes the segmentation results in supervised settings. The mechanism \textquotedblleft  AT",  \textquotedblleft ST" and \textquotedblleft SR" stand for adversarial training, self-training, and style transfer respectively.}
\label{table:segmentation_performance}
\end{table*}
We report experimental results of the proposed method on two adaptation experiments in Table\:\ref{table:segmentation_performance}. We compare our proposed method with Curriculum DA \cite{zhang2017curriculum}, CyCADA \cite{pmlr-v80-hoffman18a}, MCD \cite{saito2018maximum}, LSD-seg \cite{sankaranarayanan2018learning}, AdaptSegNet \cite{tsai2018learning}, ROAD \cite{chen2018road}, Conservative Loss \cite{zhu2018penalizing}, DCAN \cite{wu2018dcan}, and CBST \cite{zou2018unsupervised}. In Table\:\ref{table:segmentation_performance}, \textbf{Self-Ensembling (SE)} represents the segmentation performance of the network trained by source and target through the self-ensembling, without our data augmentation method. \textbf{TGCF-DA} indicates the segmentation network trained by the source data and augmented data generated from TGCF-DA with corresponding labels. \textbf{Ours (TGCF-DA + SE)} denotes our proposed framework comprised of TGCF-DA and the self-ensembling method. The proposed method significantly outperforms the baseline by 14.2\% on GTA5\(\rightarrow\)Cityscapes and 13.1\% on SYNTHIA\(\rightarrow\)Cityscapes. Our method makes further improvement compared to the source only baseline and also achieves the state-of-the-art mIoU scores on both experiments.  
    
\subsection{Ablation studies}
\begin{figure}[t]
\begin{center}
\includegraphics[width=0.7\linewidth]{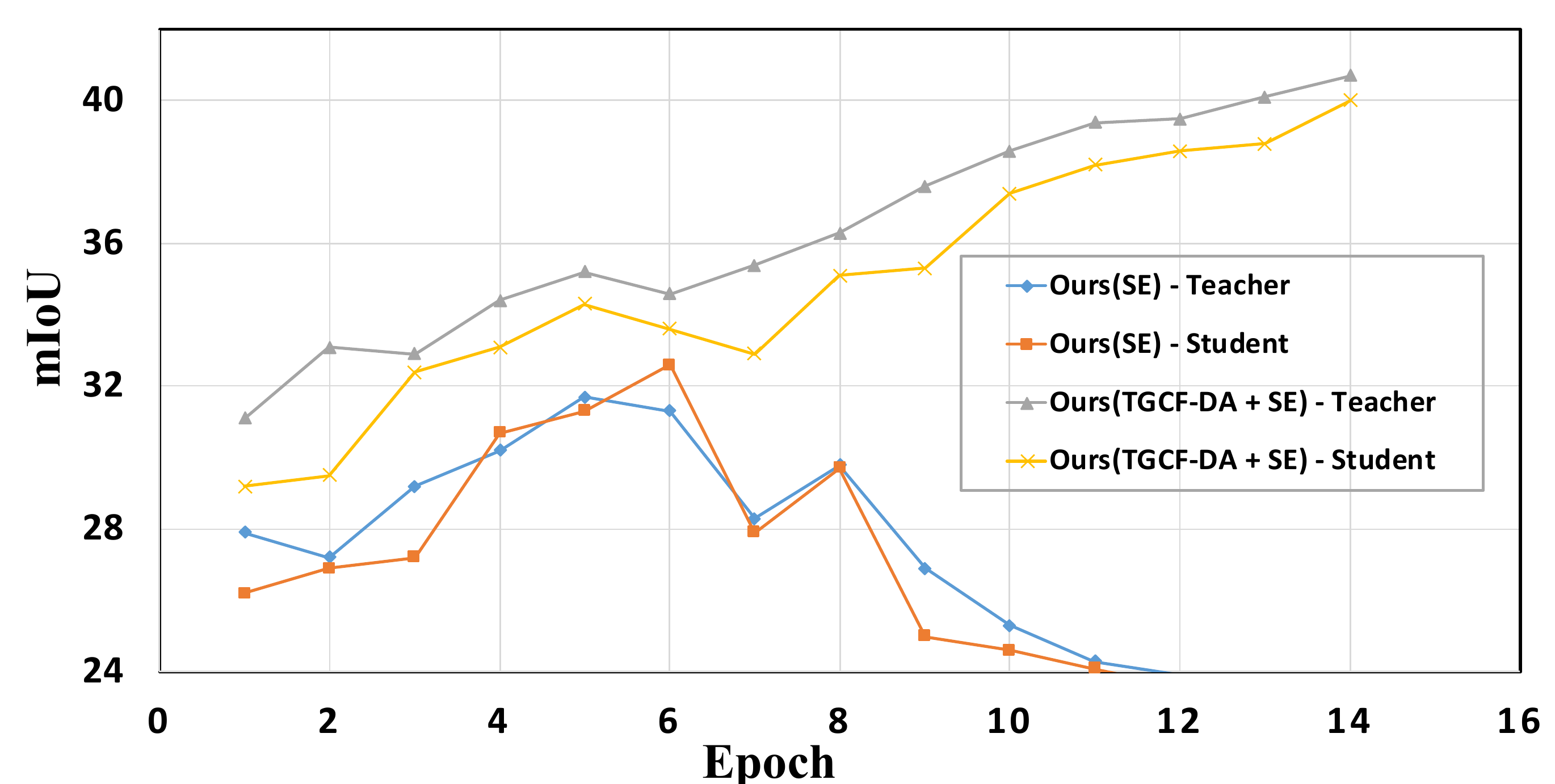}
\end{center}
   \caption{ The testing mIoUs of SE (blue and orange) and our method (grey and yellow) \(w.r.t\) training epochs on the GTA5 \(\rightarrow\) Cityscapes experiment.}
\label{fig_teacher_student_comparison}
\end{figure}

\noindent\textbf{Ablation for Self-Ensembling (SE):} Comparing the baseline and SE, SE shows small improvement in mIoUs by 4.3\% in Table \ref{table:segmentation_performance}-(a) and by 4.9\% in Table \ref{table:segmentation_performance}-(b). However, in details, we observe that SE does not perform well during the whole training process as shown in Fig.\:\ref{fig_teacher_student_comparison} (blue and orange lines). In contrast to our proposed method (TCFD-DA + SE), the teacher and student networks do not maintain complementary correlations.

\noindent\textbf{Ablation for TGCF-DA:} TGCF-DA is necessary to generate synthetic data, which help the network reduce the domain shift. Compared to the baseline, TGCF-DA improves the mIoUs by 7.1\% in Table \ref{table:segmentation_performance}-(a) and by 11.0\% in Table \ref{table:segmentation_performance}-(b). Such improvements validate that TGCF-DA serves as a useful way to reduce the domain shift. Except for TGCF-DA, SE shows the poor results in both experiments. On the contrary, our proposed method in Fig.\:\ref{fig_teacher_student_comparison} (grey and yellow lines) clearly demonstrates that the teacher updated by the student continues to improve segmentation capability, and successfully transfer its knowledge to the student. As a result, the teacher and student of our method enhance their performance simultaneously. These results substantiate our intuition that TGCF-DA enhances the capability of the self-ensembling algorithm for semantic segmentation.


\begin{figure*}[t]
\begin{center}
\centering
\includegraphics[width=0.95\linewidth]{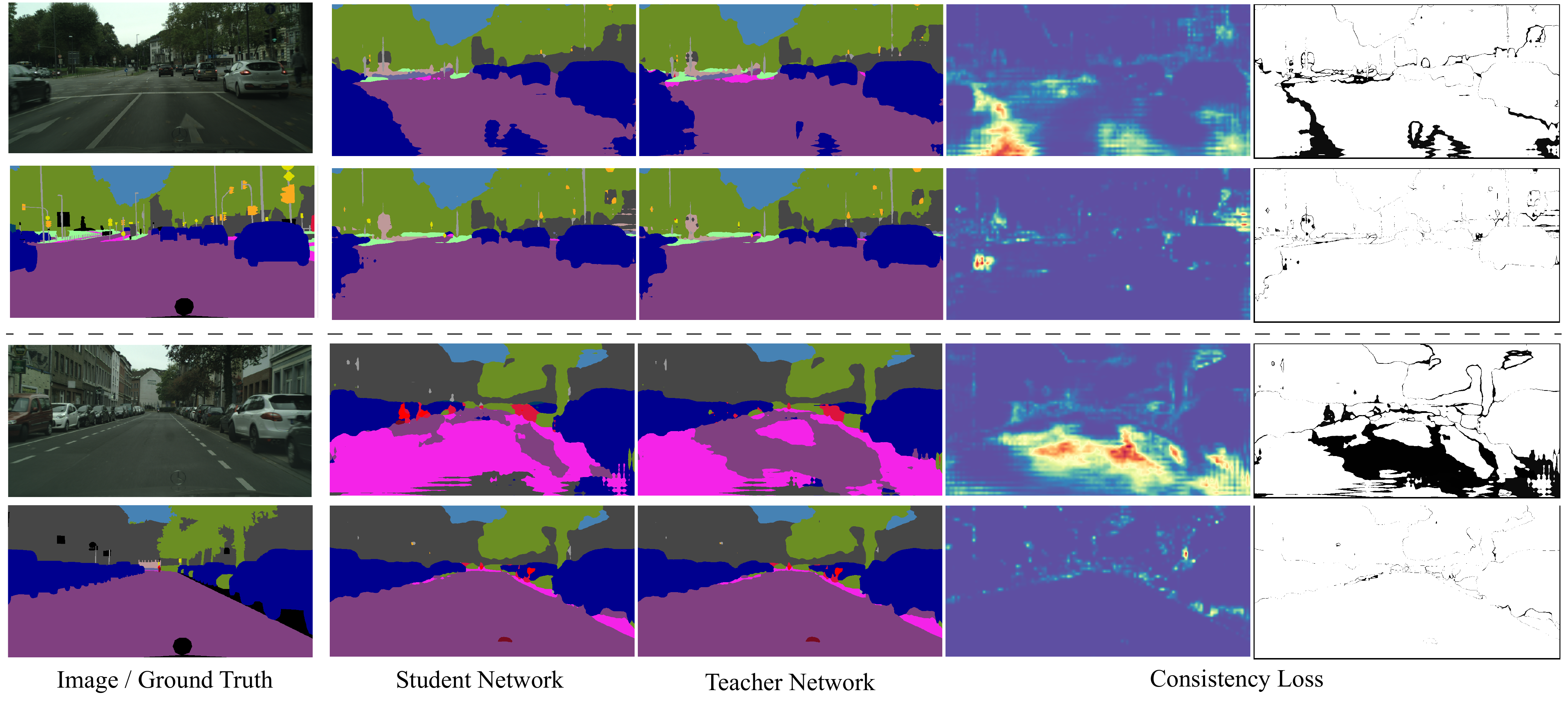}
\caption{Visualization results of GTA5 \(\rightarrow\) Cityscapes (first and second rows) and SYNTHIA \(\rightarrow\) Cityscapes (third and fourth rows). Segmentation results at 10K training steps (first and third rows) and 56K training steps (second and fourth rows). The fourth and fifth columns illustrate the heatmap of the consistency loss and disagreement map between the student and teacher networks.}
\label{fig:visualization_results}
\end{center}
\end{figure*}

\section{Analysis} \label{Section:Analysis}
In this section, we provide visualization results and analysis on varaious components of our proposed framework. 

\subsection{Visualization}
 The effectiveness of the self-ensembling is visualized in Fig.\:\ref{fig:visualization_results}. We validate that the teacher network generates better predictions, and then different predictions between the teacher and student networks cause consistency loss to enforce the consistency of their predictions. In Fig.\:\ref{fig:visualization_results}, the first and third rows show that predictions of the teacher can be a good proxy for training the student network early in the training. In addition, we point out that the consistency loss concentrates on the boundary of each object in the later training stage. Hence, the consistency loss can play a role in refining boundaries of semantic objects where the segmentation model are likely to output wrong predictions.
 
\begin{figure*}[t]
\begin{center}
\includegraphics[width=0.97\linewidth]{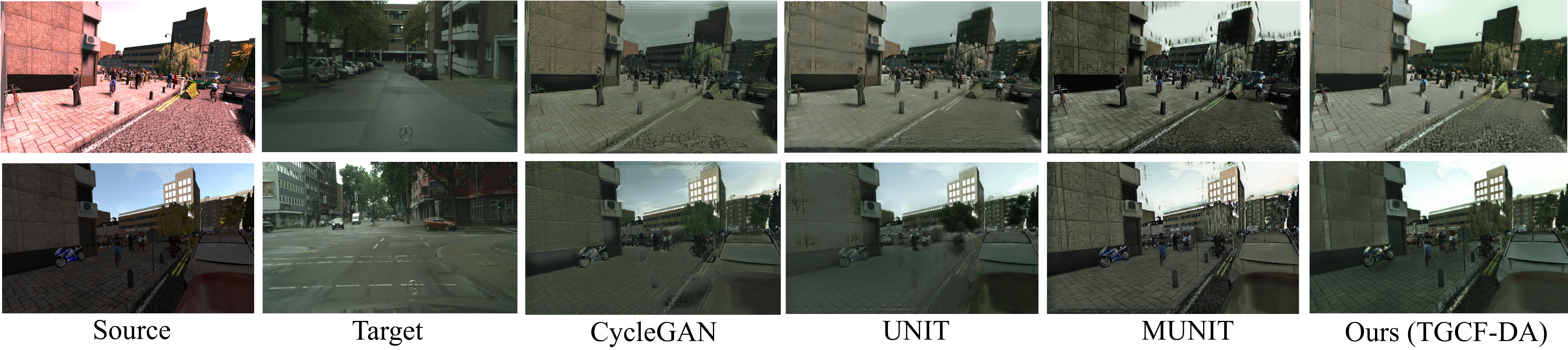}
\vspace{-3mm}
\end{center}
   \caption{Example images of SYNTHIA synthesized in the style of Cityscapes with CycleGAN \cite{zhu2017unpaired}, UNIT \cite{liu2017unsupervised}, and MUNIT \cite{huang2018multimodal}.}
\label{fig:gen_images}
\end{figure*}

In Fig.\:\ref{fig:gen_images}, we show the example results of TGCF-DA compared with other Image-to-Image (I2I) translation methods:  CycleGAN \cite{zhu2017unpaired}, UNIT \cite{liu2017unsupervised}, and MUNIT \cite{huang2018multimodal}. Both CycleGAN and UNIT often generate distorted images containing corrupted objects and artifacts. MUNIT is capable of preserving objects in images, but we observe that the style of the majority classes in the target image is often matched to elements of different classes in the source image, which is similar to \textquotedblleft spills over" problem in \cite{luan2017deep}. For example, the translated image from MUNIT shows artifacts in the sky like road texture of the target domain. Compared to the methods mentioned above, our method is not only computationally cheap and memory efficient due to the cycle-free loss but also demonstrating compelling visual results with preserving semantic consistency. 

\subsection{Analysis of self-ensembling with per-class IoUs}
\begin{figure}[t]
\begin{center}
\centering
\includegraphics[width=8cm, height=5cm]{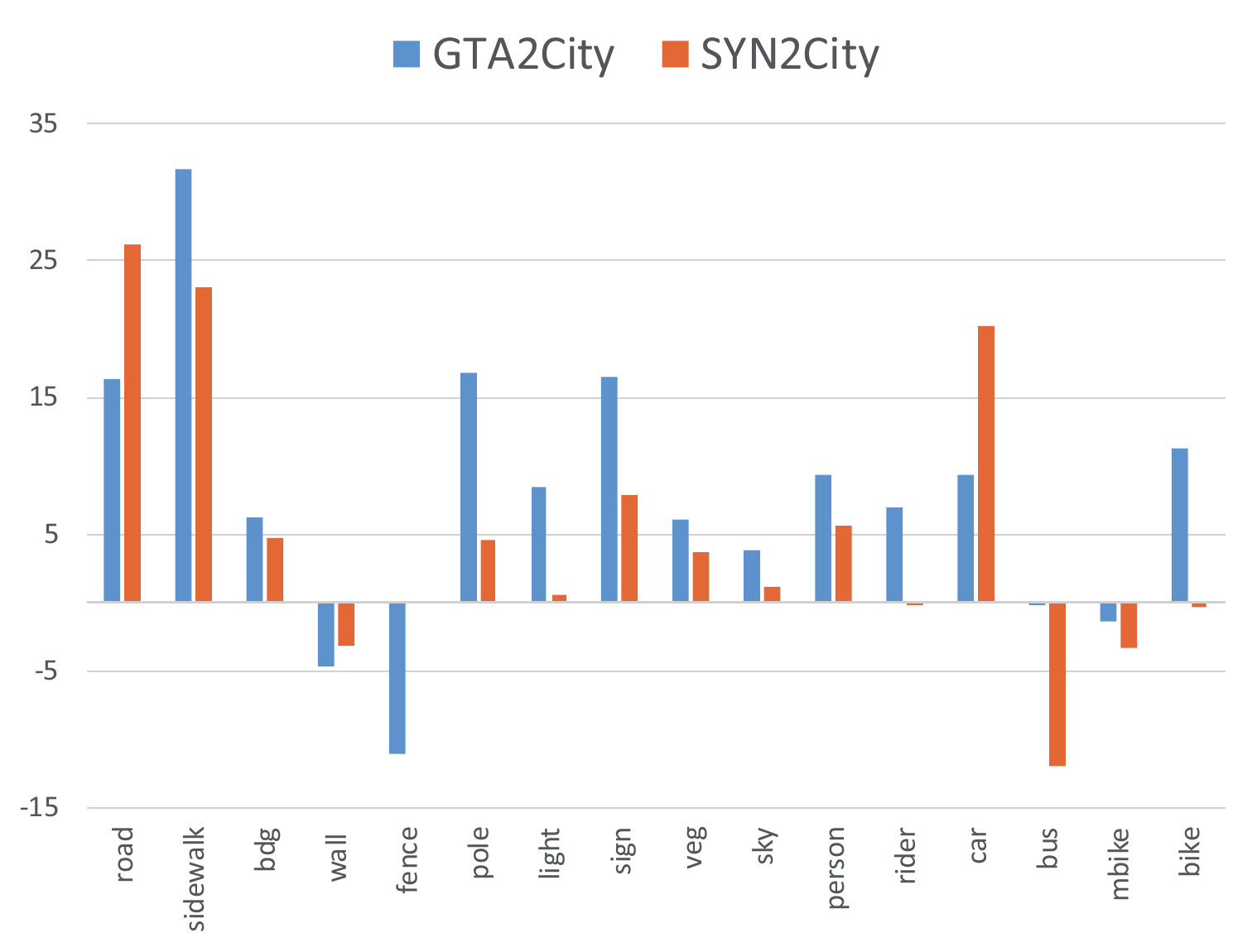}
\caption{Per-class IoU gains through the self-ensembling. The blue bar represents per-class IoU gains in GTA5\(\rightarrow\)Cityscapes experiment. The orange bar indicates the per-class IoU gains in SYNTHIA\(\rightarrow\)Cityscapes experiment.}
\label{figure:per_class_IOU_bar_graph}
\end{center}
\vspace{-6mm}
\end{figure}

To better understand the self-ensembling, we compare the per-class IoUs of our method with and without the self-ensembling. In Fig.\:\ref{figure:per_class_IOU_bar_graph}, we show the per-class IoU gains between TGCF-DA and Ours (TGCF-DA + SE). Although the IoU scores in the most categories are generally improved, there is a difference in performance gains among different categories. Figure\:\ref{figure:per_class_IOU_bar_graph} demonstrates that the IoU gains in majority classes (such as \textquotedblleft road") are generally better than those in minority classes (like \textquotedblleft bus"). These experimental results are attributed to the self-ensembling and class imbalance issues. Due to the class imbalance, the segmentation network often produces incorrect predictions on minority classes \cite{zou2018unsupervised}. In the self-ensembling method, this effect can be strengthened because the student is iteratively learned from predictions of the teacher, which tends to make incorrect predictions on minority classes rather than majority classes. Thus, the self-ensembling gives rise to large improvements in per class IoUs of majority classes compared to minority classes. It is worth noting that this result accords with our intuition that predictions of the teacher network serve as pseudo labels for the student network.

\subsection{Hyperparameter sensitivity on self-ensembling} 
\label{Subsection:Hyperparameter sensitivity on self-ensembling}
\begin{table}[t]
\begin{center}
\begin{small}
\centering
\begin{tabular}{l|c|c|c|c|c|c|c}
\toprule
\multirow{2}{*}{} &
\multicolumn{4}{c|}{Ramp-up coefficient \(\delta_{0}\)} & \multicolumn{3}{c}{EMA decay \(\alpha\)} \\
\cline{2-8}
  & 1& 3& 30& 50 & 0.9 & 0.99 & 0.999 \\
\hline
GTA5 & 41.3 & 42.3 & 42.5 & 33.6 & 37.6 & 38.9 & 42.5\\
\hline
SYN & 35.4 & 36.1 & 38.5 & 32.5 & 36.2 & 38.5 & 37.8 \\
\bottomrule
\end{tabular}
\end{small}
\end{center}
\caption{Hyperparameter sensitivity. GTA5 denotes GTA5 \(\rightarrow\) Cityscapes experiment and SYN denotes SYNTHIA \(\rightarrow\) Cityscpaes experiment.}
\label{table:Hyperparameter_sensitivity}
\end{table}
In the self-ensembling, the consistency loss weight \(\delta\) and the exponential moving average (EMA) decay \(\alpha\) are important hyperparameters. We conduct the experiments to explore the sensitivity of these hyperparameters. Table\:\ref{table:Hyperparameter_sensitivity} shows that setting a proper value for the EMA decay is significant. In all our experiments, the EMA decay is 0.99 during the first 37K iterations, and 0.999 afterward. The teacher benefits from new and accurate student's weight early in the training because the student improves its segmentation capacity rapidly. On the other hand, since the student improves slowly in the later training, the teacher can gain knowledge from the old ensembled model.          

The consistency loss weight \(\delta\) follows the formula \(\delta = 1 + \delta_{0} e^{-5(1-x)^2}\), where \(x \in [0, 1]\) denotes the ratio between the current epoch and the whole epochs and \(\delta_{0}\) is a ramp-up coefficient. Different from the usual sigmoid ramp-up \cite{tarvainen2017mean}, we add one to the formula because it is essential to guarantee the contribution of the consistency loss at the beginning of training. We decide to use \(\delta_{0}\) = 30 for all our experiments.  

\subsection{Hyperparameter sensitivity on TGCF-DA}
\begin{figure}[t]
\begin{center}
\includegraphics[width=0.99\linewidth]{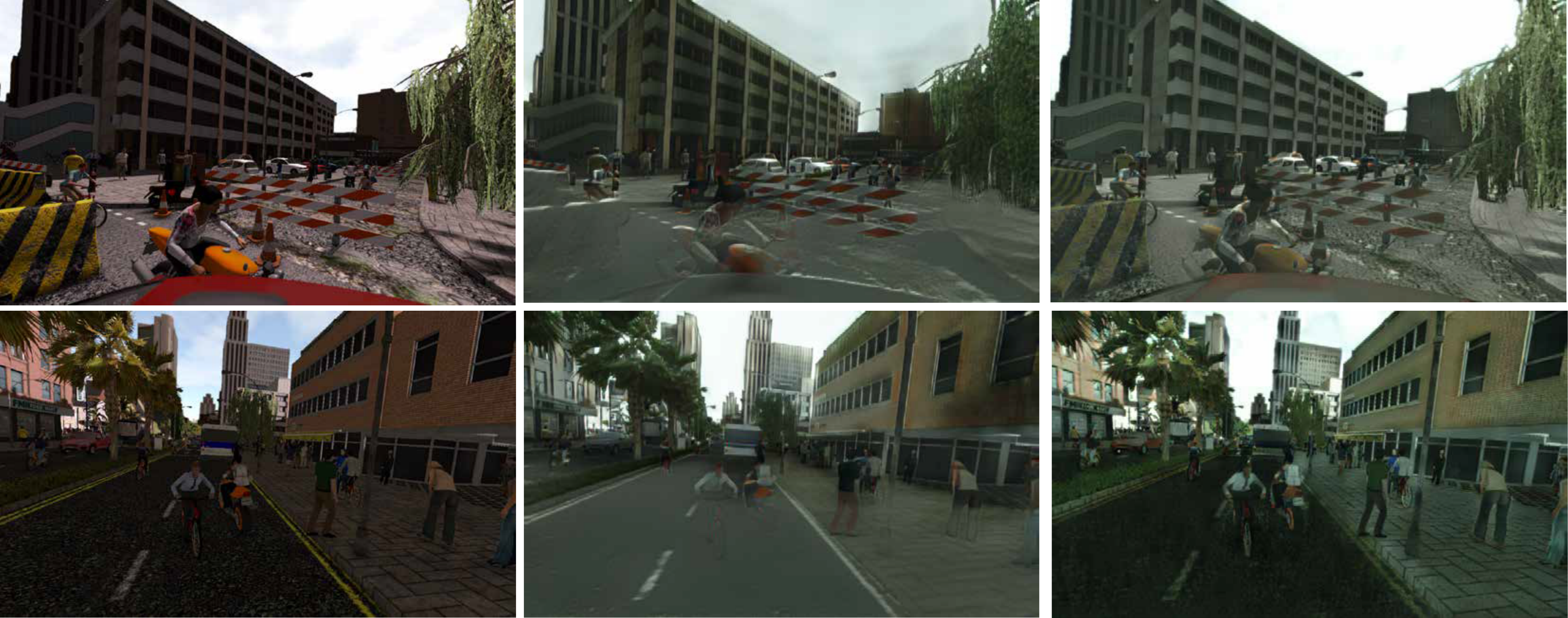}
\vspace{-2mm}
\end{center}
   \caption{The change of augmented images \(w.r.t\) the value of weight \(\lambda_{seg}\). From left to right: source input, output with \(\lambda_{seg}\) = 1, output with \(\lambda_{seg}\) = 10.}
\label{fig:hyperparam_sensitivity_on_TGCF}

\end{figure}
The weight \(\lambda_{sem}\) for the semantic constraint is a hyperparameter for training our augmentation network. Figure\:\ref{fig:hyperparam_sensitivity_on_TGCF} shows some example results on SYNTHIA \(\rightarrow\) Cityscapes. When we use a lower value (\(\lambda_{sem}\) = 1) for semantic constraint, the generator is prone to mix up objects and scenes in the augmented images. On the other hand, the proper value for semantic constraint (\(\lambda_{sem}\) = 10) helps the network preserve the local and global structures of images. These results confirm that the semantic constraint enforces our augmentation network to retain semantic consistency. 
\section{Conclusion}
We have proposed a novel framework comprised of two complementary approaches for unsupervised domain adaptation for the semantic segmentation. We present the GAN-based data augmentation with the guidance of target samples. Without the use of cycle consistency, our augmentation network produces augmented images for domain alignment. Moreover, the self-ensembling with those augmented images can perform successful adaptation by transferring pseudo labels from the teacher network to the student network. Experimental results verify that our proposed model is superior to existing state-of-the-art approaches.

\textbf{Acknowledgements} This work was supported by the National Research Foundation of Korea (NRF) grant funded by the Korea government (MSIT) (NRF-2018R1A5A7025409).

{\small
\bibliographystyle{ieee_fullname}
\bibliography{main}
}

\end{document}